\documentclass[letterpaper, 10 pt, journal, twoside]{IEEEtran} 

\IEEEoverridecommandlockouts                              

\DeclareMathAlphabet{\mathcal}{OMS}{cmsy}{m}{n}

\usepackage{graphics}
\usepackage{epsfig}
\usepackage{mathptmx}
\usepackage{times}
\usepackage{amsmath}
\usepackage{amssymb}
\usepackage{xcolor}
\usepackage[utf8]{inputenc}

\usepackage{enumitem}
\usepackage{cuted}
\usepackage{multirow}

\usepackage{mathtools}

\usepackage{subcaption}

\author{First Author$^{1}$, Second Author$^{2}$, and Third Author$^{1}$%
\thanks{Manuscript received: Month, Day, Year; Revised Month, Day, Year; Accepted Month, Day, Year.}
\thanks{This paper was recommended for publication by Editor Editor A. Name upon evaluation of the Associate Editor and Reviewers' comments.
This work was supported by (organizations/grants which supported the work.)} 
\thanks{$^{1}$First Author and Third Author are with School of Engineering, Robotics Department, University of Somewhere, Someland
        {\tt\footnotesize first.author@papercept.net}}%
\thanks{$^{2} $Second Author is with School of Engineering, Automation Department, University of Anywhere, Anyland
        {\tt\footnotesize second.author@papercept.net}}%
\thanks{Digital Object Identifier (DOI): see top of this page.}
}

\usepackage{todonotes}

\usepackage[breaklinks=true,bookmarks=false,
            pdfauthor={Saini et al.},
            pdftitle={AirPose:  Multi-View  Fusion  Network  forAerial  3D  Human  Pose  and  Shape  Estimation},
            pdfsubject={AirPose}]{hyperref}

\hypersetup{
    colorlinks=true,
    linkcolor=blue,
    filecolor=magenta,      
    urlcolor=blue,
}

\usepackage{soul}

\usepackage[font=footnotesize,labelfont=bf]{caption}

\author{Nitin Saini$^{1,2}$, Elia Bonetto$^{1,2}$, Eric Price$^{1,2}$, Aamir Ahmad$^{2,1}$, and Michael J. Black$^{1}$
\thanks{Manuscript received: August, 25, 2021; Revised December, 10, 2021; Accepted January, 4, 2022.}
\thanks{This paper was recommended for publication by Editor Pauline Pounds upon evaluation of the Associate Editor and Reviewers' comments.
The  authors  thank  the  International  Max  Planck  Research  School  for Intelligent  Systems  (IMPRS-IS)  for  supporting  Elia  Bonetto.} 
\thanks{$^{1}$First, second, third and fifth authors are with Max Planck Institute for Intelligent Systems, Tuebingen, Germany.
        {\tt\footnotesize {firstname.lastname}@tuebingen.mpg.de}}%
\thanks{$^{2}$First, second, third and fourth authors are with Institute for Flight Mechanics and Controls, The Faculty of Aerospace Engineering and Geodesy, University of Stuttgart, Stuttgart, Germany. {\tt\footnotesize {firstname.lastname}@ifr.uni-stuttgart.de}}%
\thanks{Digital Object Identifier (DOI): see top of this page.}
}

\title{AirPose: Multi-View Fusion Network for Aerial \\ 3D Human Pose and Shape Estimation}

\begin{document}

\maketitle

\begin{abstract}
   In this letter, we present a novel markerless 3D human motion capture (MoCap) system for unstructured, outdoor environments that uses a team of autonomous unmanned aerial vehicles (UAVs) with on-board RGB cameras and computation. Existing methods are limited by calibrated cameras and off-line processing.
Thus, we present the first method (AirPose) to estimate human pose and shape using images captured by multiple extrinsically {\bf uncalibrated} flying cameras. 
AirPose itself calibrates the cameras {\em relative to the person} instead of relying on any pre-calibration. 
It uses distributed neural networks running on each UAV that
communicate {\em viewpoint-independent} information with each other about the person (i.e., their 3D shape and articulated pose). 
The person's shape and pose are parameterized using the SMPL-X body model, resulting in a compact representation, that minimizes communication between the UAVs. 
The network is trained using synthetic images of realistic virtual environments, and fine-tuned on a small set of real images. We also introduce an optimization-based post-processing method (AirPose$^{+}$) for offline applications that require higher MoCap quality. 
We make our method's code and data available for research at \url{https://github.com/robot-perception-group/AirPose}. A video describing the approach and results is available at \url{https://youtu.be/xLYe1TNHsfs}.


\end{abstract}


\section{Introduction}

\IEEEPARstart{T}{hree-dimensional} human shape, pose and motion are extensively used in medical research, sports analytics, animation, gaming, etc. 
To acquire such data, several types of sensor and marker-based motion capture (MoCap) systems exist, e.g., using pressure sensors \cite{zhang2014leveraging}, reflective markers \cite{vicon} or inertial measurement units (IMUs) \cite{imu1}. 
Markerless systems using only RGB cameras are suitable for subjects that are difficult to instrument with active/passive markers or sensors (e.g., animals). 
\
Commercial markerless systems work well in controlled laboratory environments, but not outdoors in unconstrained scenarios \cite{doi:10.1080/17461391.2018.1463397}. For outdoor settings, unmanned aerial vehicles (UAVs) with on-board cameras provide an attractive solution. However, using classical MoCap techniques with UAVs present strong challenges like a markerless setting, uncontrolled lighting, complex backgrounds, and most importantly, uncalibrated, dynamic cameras.

\begin{figure}[!t]
    \centering
    \includegraphics[width=\columnwidth]{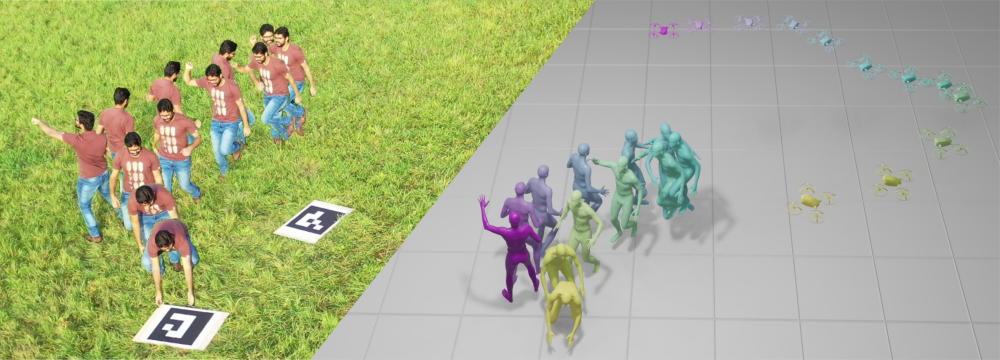}
    \caption{AirPose: A novel, distributed, multiview fusion network for 3D pose and shape estimation of humans using uncalibrated moving cameras. Real image sequence (left). AirPose$^+$ Estimates of the person (right). UAV poses are also estimated but here we manually place them for clearer illustration. ArUco markers were in the scene but never used for any of our methods.}
    \label{fig:realexpresults}
\end{figure}

Recent works like AirCap \cite{MarkerlessNitin19} use multiple UAVs to track and follow a person, record RGB images, and then post-processes them to obtain human 3D pose and shape. Thus, the UAVs cannot use human poses in realtime for their own trajectory planning. Furthermore, the self-localization estimates of the UAVs is often not accurate enough to be used for onboard camera calibration \cite{MarkerlessNitin19}. To address these challenges, in this paper we present a new method, {\em AirPose}, for on-board estimation of 3D body pose and shape of a single person from \textbf{multiple, moving, and extrinsically-uncalibrated} cameras. A novel input representation for cropped camera images allows our method to leverage monocular methods like HMR \cite{hmr} for the multi-camera scenario. The network architecture of AirPose enables distributed processing of different camera views on each UAV, and efficient sharing and fusion of relevant information between the UAVs. We develop a synthetic data generation pipeline to solve the problem of training such a network. 

Finally, we demonstrate and evaluate our approach through hardware-in-the-loop experiments using a real world-dataset collected from 2 commercial UAVs.
\
Comparison  with  existing  multi-view  methods  (incl. AirCap~\cite{MarkerlessNitin19}) is difficult because they need calibrated cameras (see Sec.~\ref{sec:sota}). A better baseline is a state-of-the-art monocular human pose estimation method, like SPIN \cite{spin}. Since SPIN (and others like it) is not trained on aerial images, we train it from scratch using our synthetic data, fine-tune it on our real datasets and use it as a baseline to compare with our method.
\
In summary, our novel contributions are: 
\textbf{(1)} A distributed and decentralized system of neural networks (AirPose) for uncalibrated moving cameras that estimates human 3D pose and shape, while simultaneously calibrating the cameras with respect to the human.
\textbf{(2)} A compact input image representation that significantly improves the human position estimate, 
even for the monocular case. 
\textbf{(3)} A realistic-looking synthetic training data generation pipeline for overhead, multi-view images of humans with ground-truth pose and shape.
\textbf{(4)} An off-board optimization-based method, AirPose${^+}$, which further refines the MoCap quality and the camera calibration.
\textbf{(5)} Code and data of our method for research purposes.

\section{Related Work}
\label{sec:sota}
\noindent \textbf{Optimization Based Methods:} State-of-the-art approach to estimate 3D human pose from multi-view images involves first extracting 2D features and then fusing them from multiple views. For example, Huang et al.~\cite{muvs} use 2D joints and silhouettes to fit the SMPL model to these features. Li et al.'s method \cite{lietal} is designed for multiple people. They detect 2D joints; use semantic segmentation to assign them to each person, and then fit SMPL to the 3D joints. 
Saini et al.~\cite{MarkerlessNitin19} propose an optimization-based method for moving cameras, where they fit SMPL to 2D joints and refine the camera parameters simultaneously. These are computationally expensive methods, unsuitable to run online on a small UAV. 
\
Schwarcz et al.'s  method \cite{schwarcz20183d} combines 2D heatmaps in each view using a conditional random field, putting constraints on temporal consistency and bone length. Amin et al.~\cite{amin2013multi} use 2D pictorial structures to compute 2D human pose and simple triangulation for 3D pose.
Pavlakos et al.~\cite{pavlakos2017harvesting} use the 3D pictorial structure model (PSM) and maximize the posterior probability of the articulated 3D pose, given the images. 
Qiu et al.~\cite{qiu2019cross} use a recursive PSM that recursively reduces discretized space bins to obtain refined 3D joints. These approaches have a speed vs performance trade-off. Finer bins give more accurate joint locations but are computationally expensive. Due to high latency and compute requirements, these are also unsuitable for deployment on UAVs.

\noindent \textbf{Deep Neural Network Based Methods:} 
Such methods employ deep neural networks to estimate 3D pose by regressing directly from the images or by combining image features given by a neural network. 
Huang et al.~\cite{Huang_2018_ECCV} train a feature extraction network that encodes the image, camera parameters, and a 3D query point together. It is slow at test time due to being query-based and does not generalize to complex backgrounds because of the prominent green background during training. 
\
Kadkhodamohammadi et al.~\cite{kadkhodamohammadi2020generalizable} train a centralized regressor for returning 3D joints using 2D joints from all views.
\
Iskakov et al.~\cite{Iskakov_2019_ICCV} perform a fusion of multiview 2D heatmaps and train a network to refine the fused volume. 
\
Remelli et al.~\cite{remelli2020lightweight} and Xie et al.~\cite{xie2020metafuse} propose networks that use camera parameters to fuse information from a pair of views to give 3D joints or improve 2D joints in each view.
\
All these methods use static and calibrated cameras, estimate only pose (not shape) and the communication overhead is high due to extensive feature sharing. 
Liang et al. address the multi-view case in \cite{Liang_2019_ICCV}, but their drawback is that the wrong information from the second view can adversely affect the output of the first even when its own input is correct. 


\noindent \textbf{Flying motion capture systems:} Flycon~\cite{flycon} and Drocap~\cite{drocap} use a single UAV with a camera. Flycon needs the subject to wear LED markers leveraging mature IR based MoCap algorithms. Drocap is a markerless system, however, it uses a high-latency fitting-based method to compute the subject's 3D skeleton and the UAV poses for the complete sequence. Flycap~\cite{flycap} uses RGB-D cameras on multiple UAVs in indoor environments to reconstruct a 3D point cloud over time. It requires a template scanning step in which the subject must be static while a UAV scans them. 
In our previous work, AirCap, we introduced autonomous formations of UAVs \cite{tallamraju2019active} to collect multiview images and optimize for 3D pose and shape offline \cite{MarkerlessNitin19} using onboard GPS-based self-localization. In the current paper, we present the first system that uses distributed neural networks for aerial MoCap, making on-board 3D body pose and shape estimation feasible, without extrinsic camera calibration.



\section{Approach}
\label{sec:methodology}

\paragraph*{Problem Statement} Our goal is to develop a method that accurately estimates the 3D pose and shape of a person from multiple uncalibrated cameras with the following constraints. It should be able to run onboard UAVs with small computation capabilities and limited wireless communication.
\
A naive approach to this problem is to use a state-of-the-art monocular human pose estimator on each UAV. This facilitates distributed and decentralized computation. However, the estimate from one UAV would not benefit from the other UAV's viewpoint. We call this the baseline method. We introduce AirPose, where information from other viewpoints is incorporated in a UAV's estimate and the computation remains distributed and decentralized.

\subsection{Baseline Method} \label{sec:baseline}
\begin{figure}[!t]
  \includegraphics[width=\columnwidth]{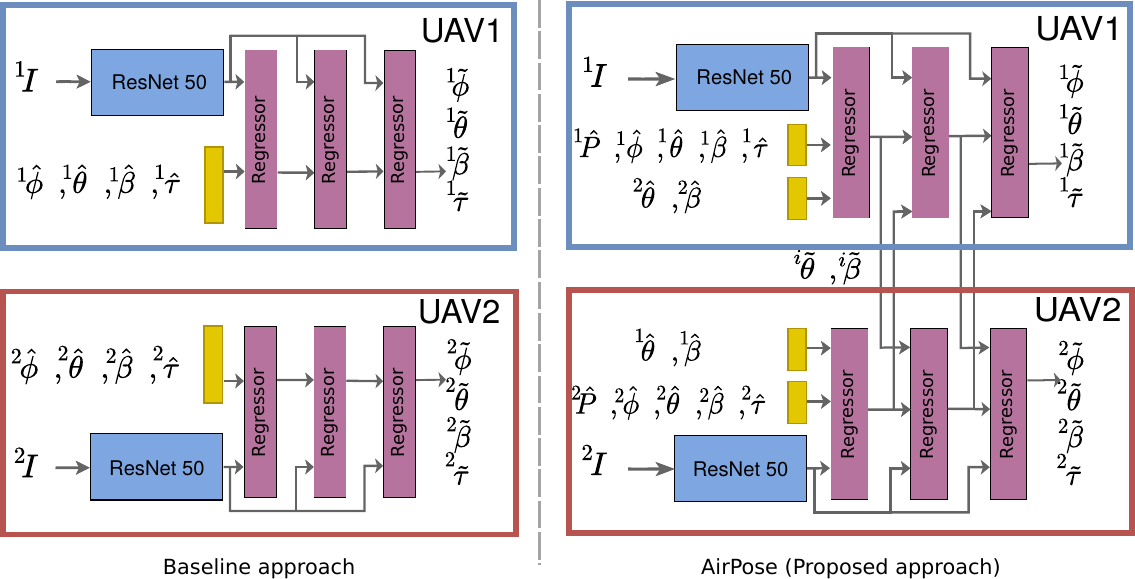}
  \caption{The network architecture of the baseline and proposed approach (AirPose). The neural network on each UAV takes in cropped and scaled image $I$ to give the body parameters relative to itself. Please refer sec. \ref{sec:methodology} for more information about the symbols.}
\label{fig:pipeline}  
\end{figure}

We adapt HMR \cite{hmr} to develop a baseline method for our problem. One network instance on each UAV outputs SMPL-X \cite{SMPL-X:2019} pose and shape estimates of the person in the UAV camera's reference frame. The setup is shown in Fig.~\ref{fig:pipeline} (left).  
\
The input consists of only a cropped and scaled region (where the person is present) of the full-size image. Thus, the output needs to be transformed to the original camera reference frame. 
\
The root translation (root refers to the root joint in the person's pose) in the original image frame is given as $\tilde{\tau}=[\tilde{x},\tilde{y},\tilde{z}]$, and in the cropped and scaled camera frame is given as $\tilde{\tau}^c=[\tilde{x}^c,\tilde{y}^c,\tilde{z}^c]$, where the vector components correspond to the 3D Euclidean coordinates. 
\
The output of the neural network is divided into four components: i) $\tilde{\tau}^c \in \mathbb{R}^3$,  ii) $\tilde{\phi} \in \mathbb{R}^6$, the root rotation, iii) $\tilde{\theta} \in \mathbb{R}^{126}$, the articulated pose, and iv) $\tilde{\beta} \in \mathbb{R}^{10}$, the body shape parameters. The dimensions of these parameters are slightly different than the original SMPL-X \cite{SMPL-X:2019} model because we use the 6D representation for rotations instead of the axis angle representation. The relationship between $\tilde{\tau}$ and $\tilde{\tau}^c$ is given as

\begin{equation}
    \tilde{z} = \tilde{z}^c s,
\end{equation}

\begin{figure}[!t]
 \includegraphics[width=0.5\textwidth]{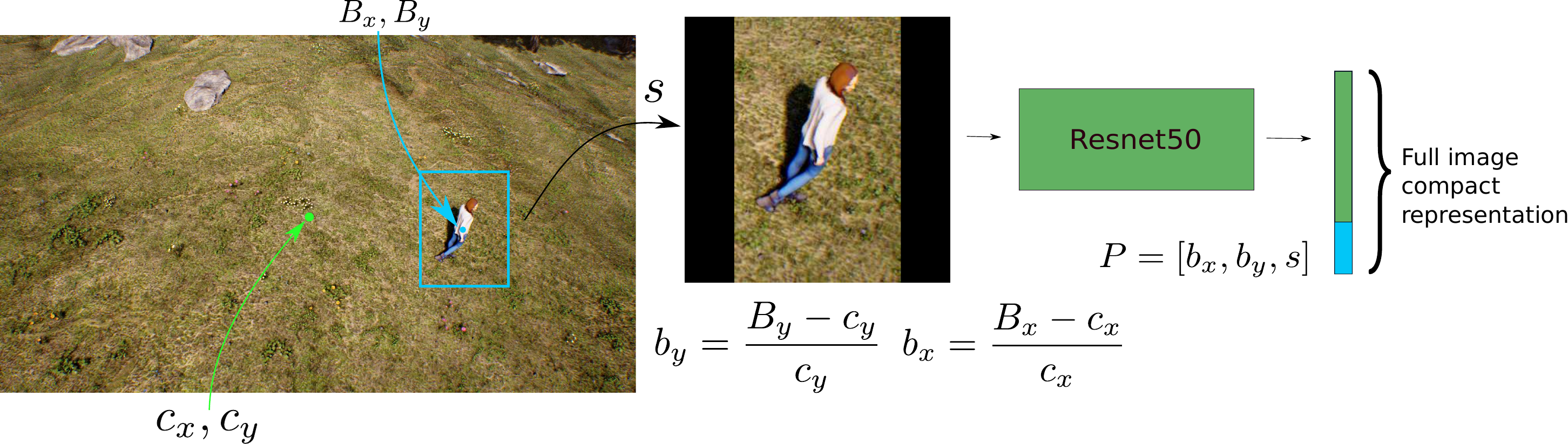}
  \caption{The bounding-box region is cropped \& scaled to the fixed size image for ResNet50 input. The full-size image is represented by concatenating the ResNet50 features and the cropping \& scaling parameter \textit{P}.}
\label{fig:resnetinput}  
\end{figure}

\begin{small}
\begin{equation}
    \begin{bmatrix}
        f_x & 0 & 0 \\
        0 & f_y & 0 \\
        0 & 0   & 1
    \end{bmatrix}
    \begin{bmatrix}
        \tilde{x}/\tilde{z}  \\ \tilde{y}/\tilde{z} \\ 1
    \end{bmatrix} = 
    \begin{bmatrix}
        f_x/s & 0 & b_x~c_x \\
        0 & f_y/s & b_y~c_y \\
        0 & 0   & 1
    \end{bmatrix}
    \begin{bmatrix}
        \tilde{x}^c/\tilde{z}^c  \\ \tilde{y}^c/\tilde{z}^c \\ 1
    \end{bmatrix},
\end{equation}
\end{small}where $f_x$ and $f_y$ are focal length parameters of the camera and ($c_x$, $c_y$) its principal point. $b_x$ and $b_y$ are the normalized coordinates of the cropped region in the original image (see Fig.~\ref{fig:resnetinput}). $s$ is the scale applied to resize the cropped region to the size $224$ by $224$, the input size of the feature extractor. As the baseline is a monocular method, we train it for one camera and run its instance independently for each UAV. The loss for our baseline method is

\vspace*{-10pt}

\begin{footnotesize}
\begin{equation}
    L_{\textrm{baseline}} = w_{j2d}L_{j2d} + w_{j3d}L_{j3d} + w_{\phi}L_{\phi} + w_{\theta}L_{\theta} + w_{\beta}L_{\beta} + w_{V}L_{V}, 
\end{equation}
\begin{equation}
    \textrm{where} ~~
    L_{j2d} = ||\mathcal{P}(\mathcal{J}(\tilde{\tau},\tilde{\phi},\tilde{\theta},\tilde{\beta})) - \mathcal{P}(\mathcal{J}(\tau_{gt},\phi_{gt},\theta_{gt},\beta_{gt}))||^2, \nonumber
\end{equation}
\begin{equation}
    L_{j3d} = ||\mathcal{J}(\tilde{\theta},\tilde{\beta}) - \mathcal{J}(\theta_{gt},\beta_{gt})||^2, ~~~ L_{\phi} = ||\tilde{\phi} - \phi_{gt}||^2,  \nonumber
\end{equation}
\begin{equation}
L_{V} = ||\mathcal{V}(\tilde{\theta},\tilde{\beta}) - \mathcal{V}(\theta_{gt},\beta_{gt})||^2, ~~~ L_{\theta} = ||\tilde{\theta} - \theta_{gt}||^2, ~ L_{\beta} = ||\beta||^2, \nonumber
\end{equation}
\begin{equation}
     w_{j2d} = 0.01, ~ w_{j3d} = 1, ~ w_{\phi} = 1, ~ w_{\theta} = 100, w_{\beta} = 1 ~ \textrm{and} ~ w_{V} = 100. \nonumber
\end{equation}
\end{footnotesize}$\mathcal{J}$ and $\mathcal{V}$ are the SMPL-X 3d joints and mesh vertices regression functions. For $L_{j3d}$ and $L_V$, the default values (zero-filled vector) are provided for $\tilde{\tau}$ and $\tilde{\phi}$. $\mathcal{P}$ is the camera projection function. $\tau_{gt}$, $\phi_{gt}$, $\theta_{gt}$, $\beta_{gt}$ are ground truth values of the corresponding SMPL-X parameters w.r.t.\ the camera. We use L2 loss function because it is shown to be efficient for these parameters in \cite{MarkerlessNitin19}\cite{SMPL-X:2019}. The loss weights in this paper are chosen as follows. Since there are many loss components, the network training is highly unstable. We first stabilize the training by selecting hyperparameters from a sparse hyperparameter space. Thereafter, we narrow down the search space. We observe that even though the overall training and validation loss keep going down, the model can overfit to some loss components. In such cases, we stop the training once the model starts overfitting on any loss component.

\subsection{Proposed Method -- AirPose} \label{sec:airpose}

We first highlight the shortcomings of the baseline method and then present our insights to solve them.

\noindent \textbf{Insight 1}  Some parts of the person's body could be occluded in one camera view, but available in other cameras. Thus, using the output of each individual network, or even simply averaging the outputs from multiple networks would not result in an accurate estimate. A systematic approach to information fusion is required to improve the estimate of the person's pose by leveraging complementary information from different views. The fusion problem is exacerbated by limited wireless communication bandwidth between the UAVs, prohibiting the realtime exchange of images among them. To perform fusion of information, while remaining within the communication constraints, we propose a novel decentralized and distributed neural network. In our proposed network, the estimated articulated pose ($\theta$) and body shape ($\beta$) from any autoregression stage of one network, running on one UAV, is fed to the next autoregression stage of another network, running on another UAV. These body parameters are independent of the viewpoints from which the person is being seen. If any body part is occluded in one view, its estimate is improved by using the information shared by the other view. There are three autoregression stages in each instance of the network, hence the total information shared per UAV per image frame is only $2\cdot(126+10) = 272$ float32.
 
\noindent \textbf{Insight 2} In the baseline method, cropping and scaling of the image result in loss of information, which is crucial for the correct estimation of the root translation. Passing the full image will result in a significant computation overhead and thus a high execution time. Our solution is to provide the network with a compact representation of the full image. This representation is the concatenation of $P$ with the extracted feature vector of the cropped and scaled image, where $P=[b_x, b_y, s]$. 

Based on the above two insights, our new network architecture is conceived as follows. It has a ResNet50 feature extractor followed by an autoregressor stage. The feature extractor extracts latent features from the cropped and scaled image similar to the baseline. These features concatenated with $P$, contain the articulated and global pose information of the person. This compact input is used by the autoregressor, along with the SMPL-X parameters, initialized as $\hat{\phi},\hat{\theta},\hat{\beta}, \hat{\tau}$ (fixed values). $\hat{\beta}$ is a vector of zeros and $\hat{\phi},\hat{\theta}$ are initialized from the same values as in \cite{hmr}. We chose the mean position of the person to be at $[0,0,10]$, normalize it by dividing with value $20$ and initialize $\hat{\tau} = [0,0,0.5]$. The normalizing factor $20$ is chosen assuming a maximum depth of $20$ meters. The estimated person position parameter $\Tilde{\tau}$ is multiplied by this normalizing factor to get the actual position.
\
The autoregressor architecture is the same as in \cite{hmr}. It consists of three fully connected (FC) layers, with a dropout layer after the first and the second FC layer. The autoregressor eventually outputs the refined SMPL-X parameters w.r.t.\ the camera. 
\
The training loss function of AirPose is

\vspace*{-10pt}

\begin{footnotesize}
\begin{equation}
    L_{\textrm{AirPose}} = w_{j2d}L_{j2d} + w_{j3d}L_{j3d} + w_{\phi}L_{\phi} + w_{\tau}L_{\tau} + w_{\theta}L_{\theta} + w_{\beta}L_{\beta} + w_{V}L_{V},
\end{equation}
\begin{equation}    
    \textrm{where} ~~
    L_{j2d} = \sum_{c}||\mathcal{P}(\mathcal{J}({}^c\tilde{\tau},{}^c\tilde{\phi},{}^c\tilde{\theta},{}^c\tilde{\beta}))\\ - \mathcal{P}(\mathcal{J}({}^c\tau_{gt},{}^c\phi_{gt},\theta_{gt},\beta_{gt}))||^2 \noindent \nonumber
\end{equation}
\begin{equation}
        L_{j3d} = \sum_{c}||\mathcal{J}({}^c\tilde{\theta},{}^c\tilde{\beta}) - \mathcal{J}(\theta_{gt},\beta_{gt})||^2, ~~~ L_{\tau} = \sum_{c}||{}^c\tilde{\tau} - {}^c\tau_{gt}||^2, \nonumber
\end{equation}
\begin{equation}
    L_{\phi} = \sum_{c}||{}^c\tilde{\phi} - {}^c\phi_{gt}||^2, ~~~ L_{\theta} = \sum_{c}||{}^c\tilde{\theta} - \theta_{gt}||^2 + ||{}^1\tilde{\theta} - {}^2\theta||^2, \nonumber
\end{equation}
\begin{equation}
    \begin{split}
    L_{V} & = \sum_{c}||\mathcal{V}({}^c\tilde{\phi},{}^c\tilde{\theta},{}^c\tilde{\beta}) - \mathcal{V}({}^c\phi_{gt},\theta_{gt},\beta_{gt})||^2
    \\
     + ||\mathcal{V}({}^1\tilde{\phi},{}^1\tilde{\theta},{}^1\tilde{\beta}) & - \mathcal{V}({}^2\phi,{}^2\theta,{}^2\beta)||^2, ~~ L_{\beta} = \sum_{c}||{}^c\beta||^2 + ||{}^1\beta - {}^2\beta||^2, \nonumber
    \end{split}
\end{equation}
\begin{equation}
    w_{j2d} = 0.002, ~ w_{j3d} = 1, ~ w_{\phi} = 1, w_{\tau} = 10, ~ w_{\theta} = 50, ~ w_{\beta} = 1 ~ \textrm{and} ~ w_{V} = 50.\nonumber
\end{equation}
\end{footnotesize}Each camera has its own SMPL-X parameter estimates, e.g., ${}^c\theta$ is the articulated pose parameter for camera $c$, $c \in \{1,2\}$.

\smallskip

\noindent \textbf{Insight 3} Most monocular methods are trained on data that does not contain overhead and oblique views of persons. In aerial MoCap, such viewpoints are predominant. On the other hand, there also exist few multiview image dataset on which our baseline method could be trained, let alone those with overhead viewpoints. To address both these dataset related challenges, we train our networks (both baseline and AirPose) using large sets of synthetic images in realistic virtual environments and fine-tune  using a small set of real images from UAVs. During the finetuning, the weights of ResNet50 feature extractors are frozen and only the regressor is trained. We use OpenPose \cite{openpose1} to get the 2D keypoints on the images and use them for the supervision during the fine-tuning. To make the 2D keypoints more reliable, we also get them from another detector, AlphaPose \cite{fang2017rmpe}. If the OpenPose estimate deviates from the AlphaPose estimates by more than a threshold value (of 100 pixels), it is discarded. Using only the 2D keypoints for supervision might result in unnatural body poses. Thus, we use a variational autoencoder network VPoser \cite{SMPL-X:2019}, a learned prior distribution of human body poses using a large dataset of human poses. We use its encoder network ($\mathcal{E}$) to project the estimated poses from our network into the latent space and restrict it to be close to the mean of the VPoser's distribution. The fine-tuning loss for the baseline approach is given as

\begin{footnotesize}\begin{equation}
    L_{f_\textrm{Baseline}} = w_{j2d}L_{j2d} + w_{\beta}L_{\beta} + w_{vposer}L_{vposer}, 
    \label{eqn:baseline_real}
\end{equation}
\begin{equation}
    \textrm{where} ~~
    L_{j2d} = \sum_j{w_{j}((\mathcal{P}(\mathcal{J}(\tilde{\tau},\tilde{\phi},\tilde{\theta},\tilde{\beta})) - J_{j})^2)}, ~~~ L_{\beta} = ||\beta||^2,  \nonumber
\end{equation} %
\begin{equation}
    L_{vposer} = ||\mathcal{E}(\theta)||^2, ~~~ w_{j2d} = 0.01, ~~~ w_{\beta} = 5 ~~~ \textrm{and} ~~~ w_{vposer} = 1. \nonumber
\end{equation}\end{footnotesize}$J_{j}$s are the 2D coordinates estimated by OpenPose for keypoint $j$. $w_{j}$s are the confidence scores of the corresponding keypoint estimates from OpenPose. $\mathcal{E}(\theta)$ is the sample from the latent space distribution obtained after passing $\theta$ to the VPoser encoder. The fine tuning loss for AirPose is

\begin{footnotesize}
    \begin{equation}
        L = w_{j2d}L_{j2d} + w_{\beta}L_{\beta} + w_{vposer}L_{vposer} + w_{\theta}L_{\theta},
    \label{eqn:airpose_real}
    \end{equation}
    \begin{equation}
        \textrm{where} ~~ L_{j2d} = \sum_{j,c}{{}^cw_{j}((\mathcal{P}(\mathcal{J}({}^c\tilde{\tau},{}^c\tilde{\phi},{}^c\tilde{\theta},{}^c\tilde{\beta})) - J_{j})^2)}, ~~~ L_{\theta} = ||{}^1\theta - {}^2\theta||^2 \nonumber
    \end{equation}%
    \begin{equation}
        L_{\beta} = \sum_{c}(||{}^c\beta||^2) + ||{}^1\beta - {}^2\beta||^2, ~~~ L_{vposer} = ||\mathcal{E}(\theta)||^2 \nonumber
    \end{equation}
    \begin{equation}
        w_{j2d} = 0.01, ~~~ w_{\beta}=5, ~~~ w_{vposer} = 0.1 ~~~ \textrm{and} ~~~ w_{\theta}=100.
    \end{equation}
    \end{footnotesize}
    
\subsection{Proposed Approach -- AirPose$^+$} \label{sec:airpose+}
Finally, we propose a post-processing optimization method, AirPose${^+}$, where we utilize the temporal information to further refine the human pose and shape and camera pose estimates given by AirPose. This is done by minimizing the loss function given in (\ref{eqn:airpose+loss}). We optimize the parameters $\theta$, $\beta$, ${}^c\phi$, ${}^c\tau$ for the whole capture sequence. We also put a constraint over these parameters to be close in adjacent frames. We use the latent representation of VPoser $v$ to represent articulated pose. The SMPL-X pose parameter $\theta$ can be obtained by passing $v$ through the vposer decoder $\mathcal{D}$ i.e. $\theta$ = $\mathcal{D}(v)$.

\begin{footnotesize}
        \begin{equation}\label{eqn:airpose+loss}
            L = \sum_t (w_{j2d}L_{j2d} + w_{vposer}L_{vposer} + w_{temp}L_{temp}) + w_{\beta}L_{\beta},
        \end{equation}
        \begin{equation}
            \textrm{where} ~~ L_{j2d} = \sum_{j,c}{{}^cw_{j}(\rho(\mathcal{P}(\mathcal{J}({}^c\tilde{\tau},{}^c\tilde{\phi},\mathcal{D}(v),\tilde{\beta})) - J_{j}))}, ~~~ L_{vposer} = ||v||^2, \nonumber
        \end{equation}
        \begin{equation}
              L_{temp} = ||{}_t\tilde{\theta}-{}_{t-1}\tilde{\theta}||^2 + \sum_c ||{}_t\tilde{\phi}-{}_{t-1}\tilde{\phi}||^2 + ||{}_t\tilde{\tau}-{}_{t-1}\tilde{\tau}||^2, \nonumber
        \end{equation}
        \begin{equation}
             L_{\beta} = ||\tilde{\beta}||^2, ~~~ w_{j2d} = 1, ~ w_{vposer} = 0.05, ~ w_{temp} = 1 ~ \textrm{and} ~ w_{\beta} = 2000. \nonumber
        \end{equation}    
    \end{footnotesize}$\rho$ is the Geman-Mcclure robust penalty function. All the individual loss terms (except $L_{\beta}$) in (\ref{eqn:airpose+loss}) are function of $t$. However, $t$ is not used in their notation to improve the readability. Unlike AirPose, in AirPose$^+$, articulated pose $v$ is not different for each camera and $\beta$ is constant for each camera throughout the sequence. We calculate $\sum_{j,c}{}^cw_j$ for each frame $t$ and ignore it if its value is below a threshold.


\section{Training and Evaluation}
\label{sec:training}
An ideal dataset for training our network requires synchronized video sequences of many persons with a variety of poses and with ground truth SMPL-X parameters. It is difficult to collect such data because of reasons like limited battery life of UAVs, weather-related uncertainties, flying permission from authorities, availability of licensed pilots etc. Thus, we generate realistic-looking synthetic image data in realistic virtual environments. We also collect a smaller amount of real data with two persons and two DJI UAVs (7000 frames per person per UAV). We fine-tune our network using the real data with one person and evaluate it on the other person. 

\smallskip

\textbf{Synthetic Data for Training:} We generated the synthetic data ($\sim$ 30000 frames per UAV) by putting realistic human scans \cite{agora} in Unreal Engine (UE) and render them from multiple viewpoints of 2 UAVs. We use AirSim plugin \cite{airsim2017fsr} for UE to move the scans and the cameras around such that the data generation process is automated. The scans are put in an outdoor UE environment (purchased from UE Marketplace) and moved between -2.75m to 2.75m along the X and the Z-axis (the Y-axis points outward from the ground plane in UE). The cameras are moved independently and randomly around the origin such that they are facing the origin and the distance from the origin is $\sim$10m. The pitch of the cameras varies from $0^{\circ}$ to $45^{\circ}$. The SMPL-X fittings to scans are provided by \cite{agora}. The fittings are done using the gender-specific SMPL-X model.


\smallskip

\textbf{Real Data for Fine-tuning:} We collected 2 real data sequences. Each sequence uses two DJI Mavic UAVs, flying manually around a different person. The raw data are sets of video sequences from the UAVs. During the acquisition, we keep one of the UAVs hovering in place and the other one is manually flown around the person, who performs various motions, covering a wide range of poses at a safe distance from the UAVs. The intrinsic calibration of the UAV cameras was done immediately before the take-off, using the chessboard calibration method. The frames were extracted and manually time-synchronized. We sampled the corresponding frames from the two UAVs and found that they had a constant time difference, which implies that the two devices had the same frame rate and it remained constant throughout the acquisition time. ArUco markers were in the scene but never used for any of our methods.

\textbf{Hardware-in-the-loop evaluation and synchronization strategy for online execution:}
In order to run our approach on-line we took a hardware-in-the-loop approach, where communication and synchronisation were performed in real-time on our actual UAV hardware (which has intel I7 CPU and Nvidia Jetson TX2 on-board) \cite{tallamraju2019active}. We implemented a ROS~\cite{ros} based synchronization framework that selects matching camera frames based on their shutter timestamp. Clocks were synchronized using NTP~\cite{ntp1991}. Communication via 5 GHz Wifi with 50 Mbps was handled by ROS. On the Jetson TX2 modules, AirPose neural network achieved inference speeds of 50 ms per frame (224x224x3 bytes 20 fps). This was dominated by the first step of the network (i.e. ResNet50 and the first regressor stage, $\sim43 $ ms) while the other two steps are almost immediate ($\sim2.5$ ms). However, in a real world setting high-resolution images and communication must be accounted for. 
\
We allocated 2 * 25 ms for the communication of the 544 byte large encoding, as well as 140ms for acquisition and downsampling of the 4K camera images after benchmarking these operations individually on our UAVs. 
\
To accommodate sufficient time for image acquisition, inference, and multiple iterations of data communication, we allocated a fixed length time-window of 240 ms to process one frame, which results in a fixed overall framerate of 4.17 fps.
\
If processing could not complete by the end of the window, the frame was discarded. Taking dropped frames into account AirPose achieved an average framerate of around 3 fps. The camera itself was capturing 4K frames at 40 fps. It must be noted that the bottleneck in this ad-hoc approach is not the AirPose neural network, but acquisition and preprocessing as well as communication overhead. If lower resolution camera images were to be used in conjunction with optimized buffers to acquire, process and communicate data in parallel, it is easily conceivable that AirPose could reach 20fps realtime-throughput at a latency of less than 150ms.
Real image datasets from~\cite{tallamraju2019active, DeepPrice18} are not directly usable in our approach due to significant differences with respect to the appearance of the images on which the network is originally trained (i.e. synthetic data). To overcome this problem, we collected a custom video dataset with two DJI drones for evaluation (similar to the data for fine tuning) and converted them into rosbags. We manually synchronized the first common frame between the two video sequences. Subsequently, we replayed the rosbags as if they were taken by the UAV cameras and used the hardware-in-the-loop setting and synchronization architecture as explained above.


\section{Experiments and Results}
\label{sec:exp_results}
\subsection{Results on the Synthetic Data}
\label{sec:synth_results}
\begin{figure*}[!ht]
\setlength{\belowcaptionskip}{-10pt} 
\hspace{5pt}
\includegraphics[width=0.9\textwidth]{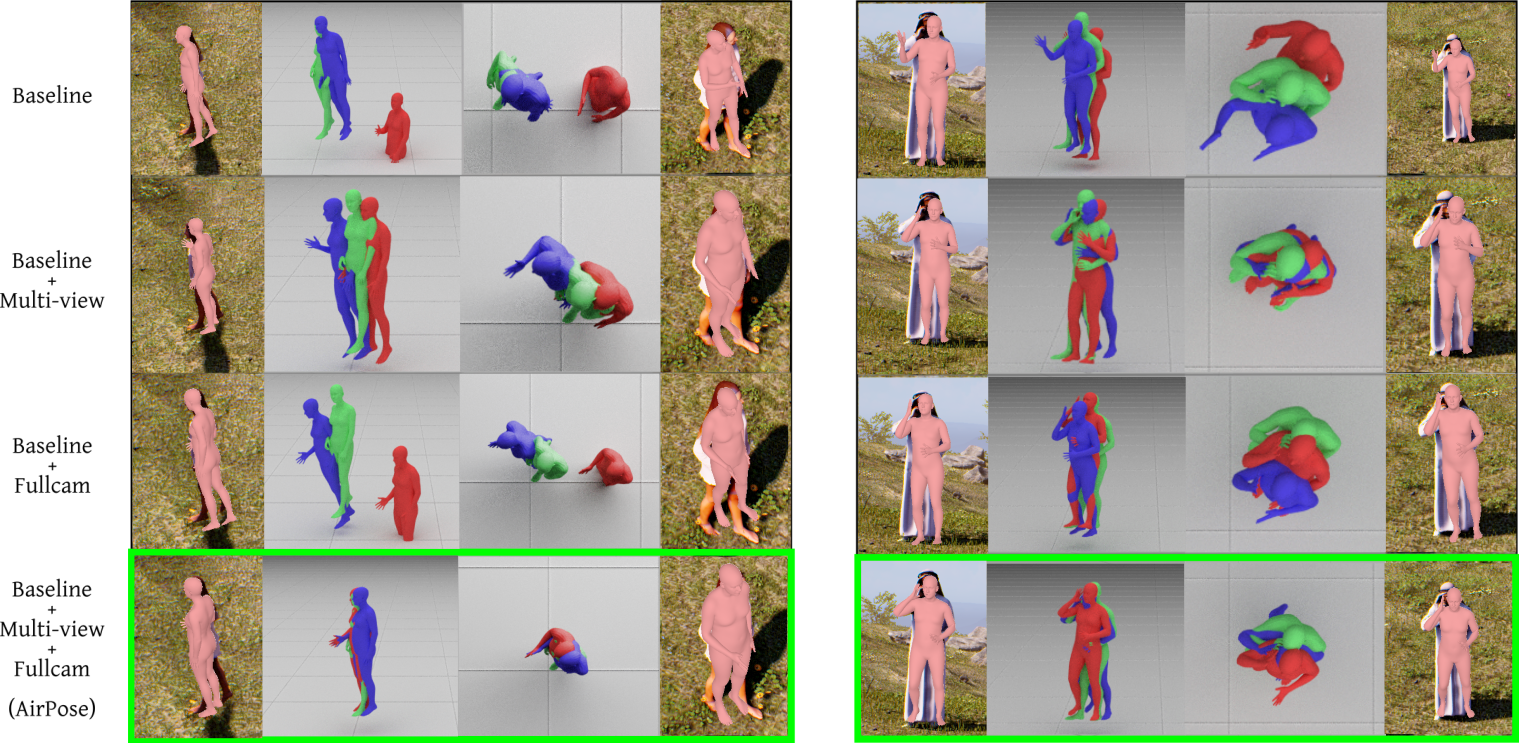}
\caption{Ablation Study (Sec.~\ref{sec:synth_results}) qualitative results. First row: `Baseline'. Row 2: `Baseline+Multi-view'. Row 3: `Baseline+Fullcam'. Row 4: `AirPose'. 1$^{st}$ and 4$^{th}$ columns are cropped images showing the overlaid estimated mesh w.r.t.\ the camera. The 2$^{nd}$ and 3$^{rd}$ columns show the front and the top views of the 3D scene, where the two estimates (one from each view) are transformed to the global coordinate frame. Red mesh: estimate from the first camera. Blue mesh: estimate from the second camera. Green mesh: ground truth SMPL-X mesh.}
\label{fig:fig_synthetic_results_visual}
\end{figure*}
\begin{figure*}[!htbp]
  \includegraphics[width=0.5\textwidth]{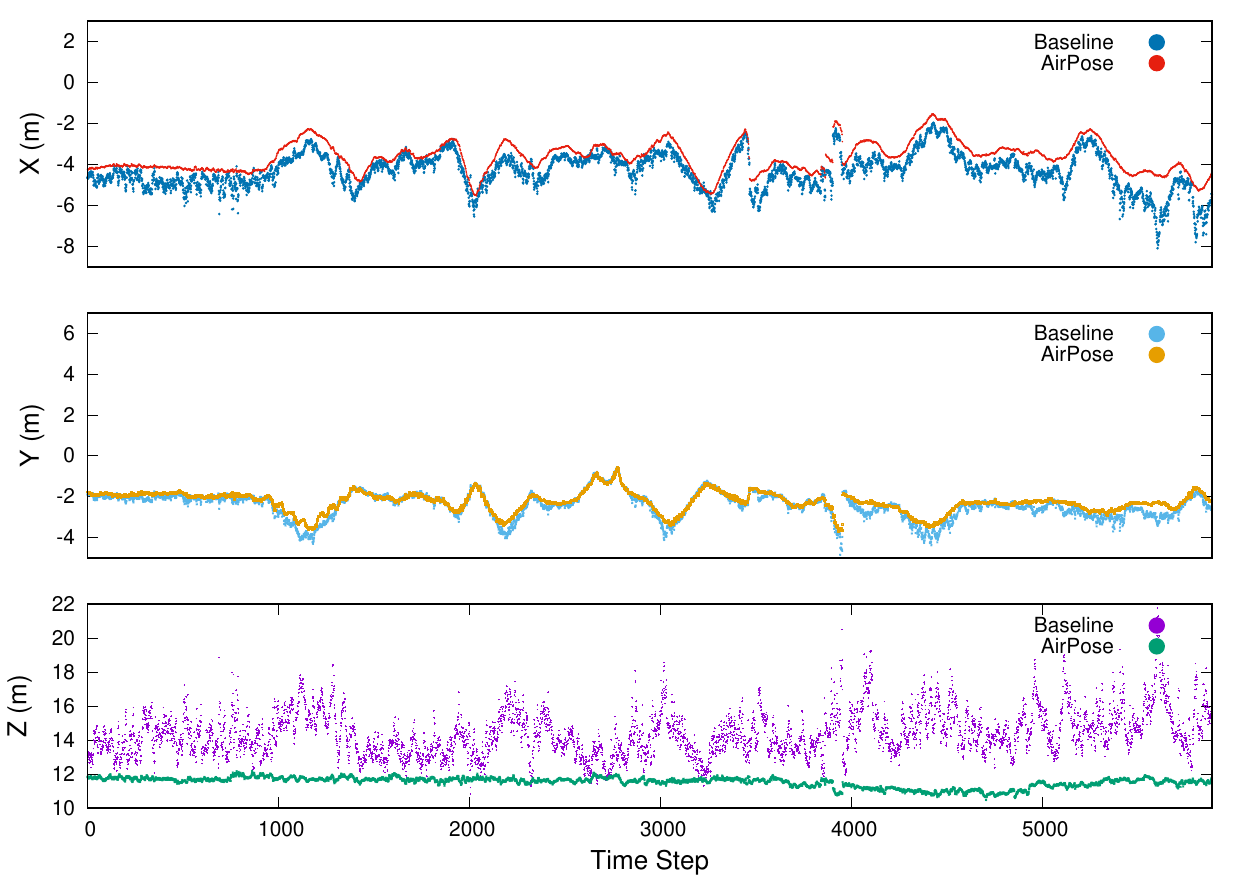}
  \includegraphics[width=0.5\textwidth]{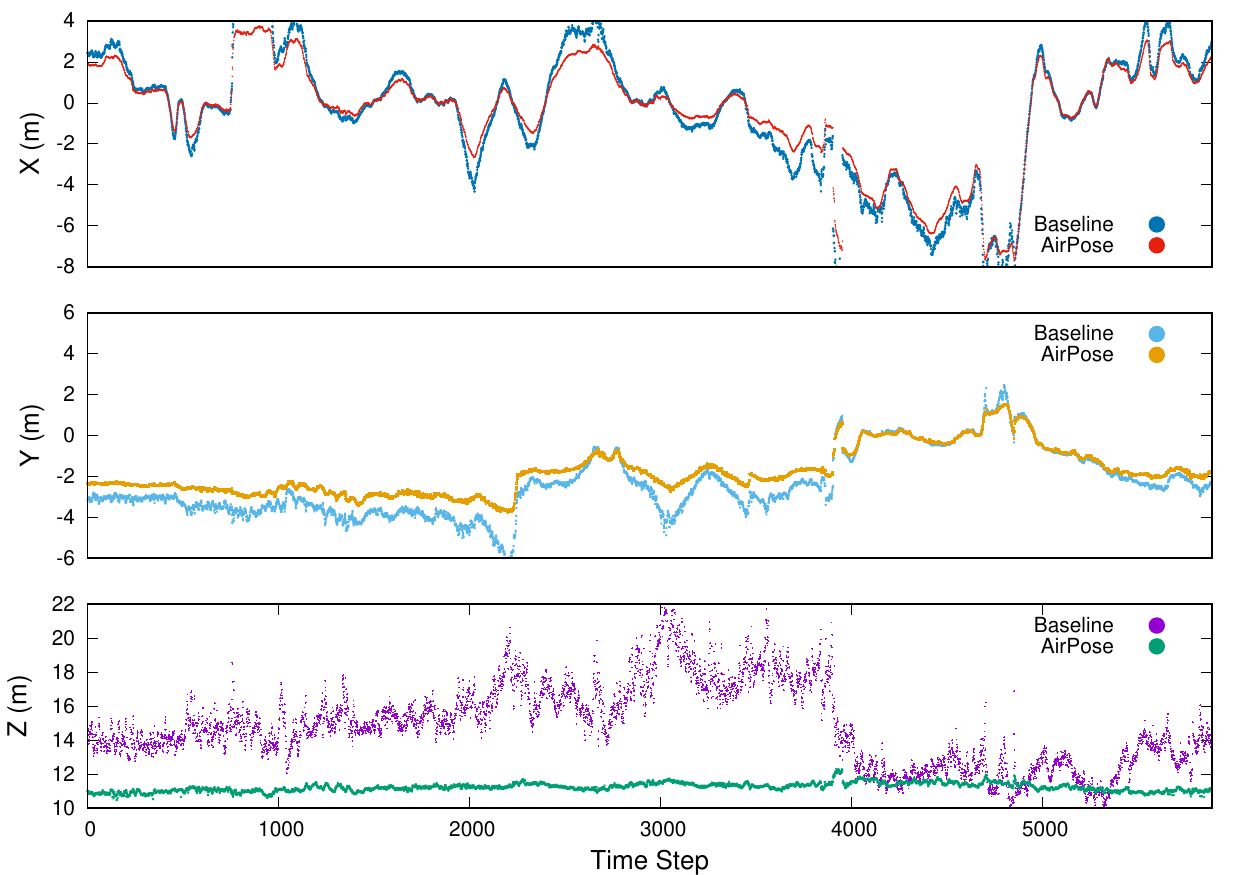}
  \caption{Comparison of the person 2 3D position trajectory, estimated by baseline and AirPose methods on UAV 1 (hovering, left) and 2 (circling, right).}
\label{fig:cam1_trajectory}  
\end{figure*}

We perform an ablation study and compare the following 4 methods using the synthetic data.

\smallskip
  
\noindent \textbf{1. Baseline}: The baseline method as described in Sec.~\ref{sec:baseline}.

  \smallskip
    
\noindent \textbf{2. Baseline + Multi-view}: Here we expand the input size of the baseline regressor to take the $\theta$ and $\beta$ values from another view. However, here the regressor does not take the image cropping and scaling information $P$ and estimates the SMPL-X parameters w.r.t.\ the cropped image. 
    
    \smallskip
    
\noindent \textbf{3. Baseline + Fullcam}: Here, in addition to the baseline inputs, we take the cropping and scaling information, $P$, and estimate the pose of the person w.r.t.\ the original camera. It is equivalent to AirPose without communication between the two UAVs.
  
  \smallskip
  
\noindent \textbf{4. AirPose (Baseline + Multi-view + Fullcam)}: The proposed method as described in Sec.~\ref{sec:airpose}.

\smallskip


\begin{table}[!b]
    \centering
    \begin{footnotesize}
    \begin{tabular}{ p{1.4cm} | c | p{1.4cm} | p{1.4cm} | c }
        & Baseline & Baseline + Multi-view & Baseline + Fullcam & AirPose \\
        \hline 
      MPE (m) & 0.50 & 0.46 & 0.22 & \textbf{0.15} \\ \hline
      MPJPE (m) & 0.091 & 0.084 & 0.077 & \textbf{0.072}    
    \end{tabular}
    \end{footnotesize}
    \caption{{\bf Ablation study} of AirPose on the synthetic dataset.}
    \label{tab:ablation}
\end{table}

\noindent \textbf{Error Comparison Metrics:} We compare the global position and articulated pose estimates of the person from the four methods. Since, the person pose is estimated by each UAV relative to itself, for computing error we convert the estimates in the global frame using the ground truth extrinsics of the UAVs. We calculate the error of each estimate w.r.t.\ the person's ground truth (GT) in the global frame. To evaluate the global position estimate, we calculate the mean position error (MPE) as $\textrm{MPE} = \frac{1}{2N}\sum_{n=1}^N\sum_{c}||{}^c\tilde{\tau}^o - \tau_{gt}^o||$. It is the mean (over all the images from both UAVs) of the Euclidean distances between the GT and the estimated person positions in the global frame (denoted by the superscript ${}^o$). 



    

For the evaluation of the articulated pose estimate of the person, we calculate the mean of the joint position errors (MPJPE) as $\textrm{MPJPE} = \frac{1}{22(2N)}\sum_{n=1}^N\sum_{j}\sum_{c}||{}^c\tilde{\tau}^o - \tau_{gt}^o||$, over all the images from both UAV cameras. $\tau$ is a function of $j$ but it is not denoted here to improve the readability. The joint error is the Euclidean distance between the estimated joint position and its corresponding GT when the root translation for both the estimate and GT are aligned. The number of joints in our case is 22.
%

\begin{figure}[!htp]
    \centering
    \includegraphics[trim=70 0 80 0, clip, width=0.48\textwidth]{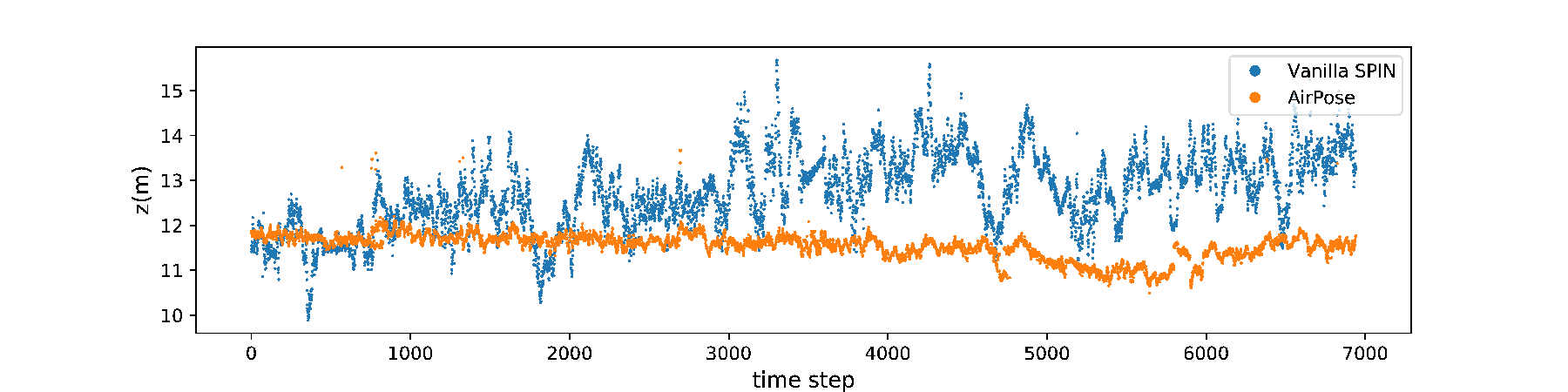}
    \caption{Comparison of the estimated Z coordinate of the tracked person's trajectory by AirPose and the vanilla SPIN.}
    \label{fig:spinRes}
\end{figure}

\noindent \textbf{Results and Discussion:} Table \ref{tab:ablation} shows the MPE and MPJPE values for the four different methods. 
AirPose, outperforms all ablated methods on both the metrics. 
The MPE of `Baseline' and `Baseline + Multi-view' is similar and much higher than the other two methods. This highlights the problem of using cropped images without the full image information. 
The input image does not contain information about the position of the person in the full image, and the camera center is incorrectly assumed to be the center of the bounding box. 
\
Since the person is viewed from above, there are less self-occlusions present in the dataset. Due to this, all the methods have similar MPJPE. Nevertheless, AirPose, combining information from both the views, has the least MPJPE.
\
In summary, the ablation shows that our insights and the proposed solutions (in Sec.~\ref{sec:methodology}) are critical: 
information about the position of the person in the 2D image improves position estimates in 3D,  and exploiting information from the other view substantially improves articulated pose estimates. Our proposed method utilizes both of these factors and thus results in a significant improvement of both the MPE and MPJPE.

Figure~\ref{fig:fig_synthetic_results_visual} provides a qualitative analysis of the ablation study.
Note that, from the camera view, the results of the different methods do not appear to differ significantly; human pose estimates projected into the camera can be misleading.
In contrast, the 3D views clearly illustrate the errors in pose and shape.

\begin{figure*}[!htbp]
    \includegraphics[width=\textwidth]{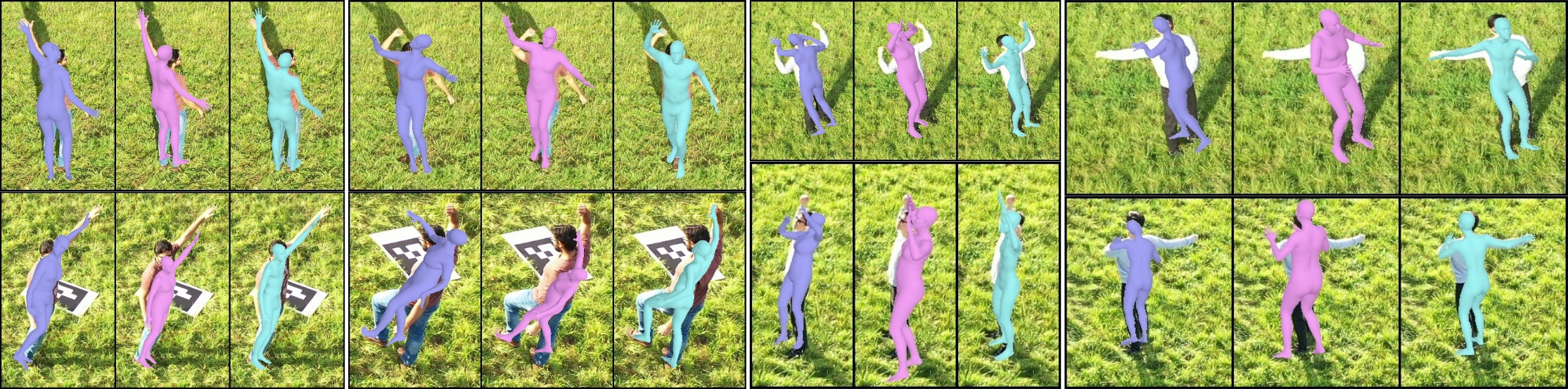}
    \caption{{\bf Real data.} Baseline, AirPose and AirPose$^+$ results on 4 real data samples. Each sample contains a 2x3 grid of images with the estimated mesh overlaid. Each row corresponds to one camera view and each column shows projected results from the baseline (blue), AirPose (pink) and AirPose$^+$~(cyan). The first two samples show person 1 and the remainder show person 2.}
  \label{fig:real_data_results}  
  \end{figure*}



\subsection{Results on the Real Data}

\begin{figure*}[!htbp]
    \setlength{\belowcaptionskip}{-10pt} 
    \setlength{\abovecaptionskip}{-5pt}  
  \begin{tabular}{cc}
  \hspace*{-20pt} \includegraphics[width=0.53\linewidth]{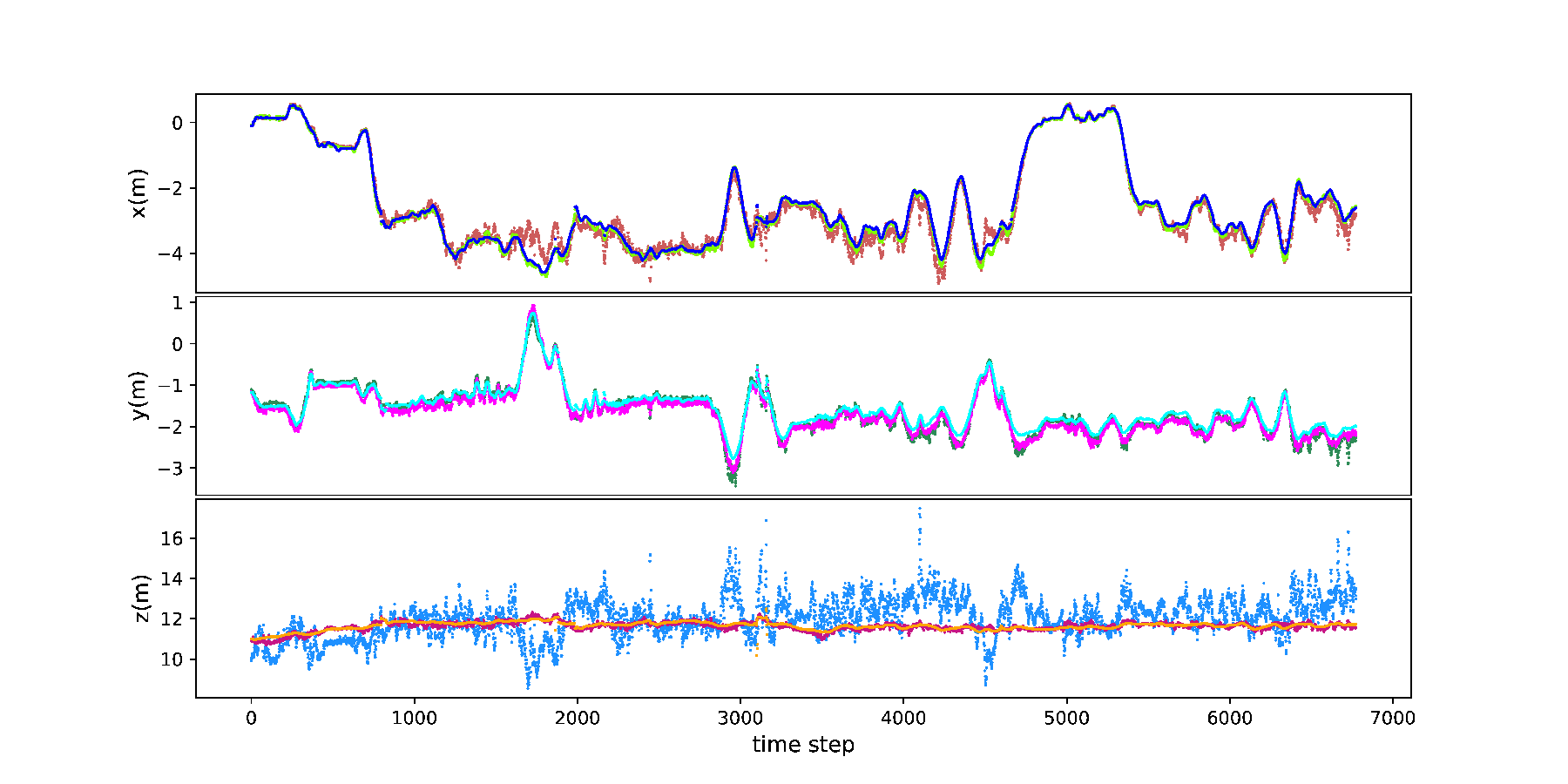} & \hspace*{-35pt}\includegraphics[width=0.53\linewidth]{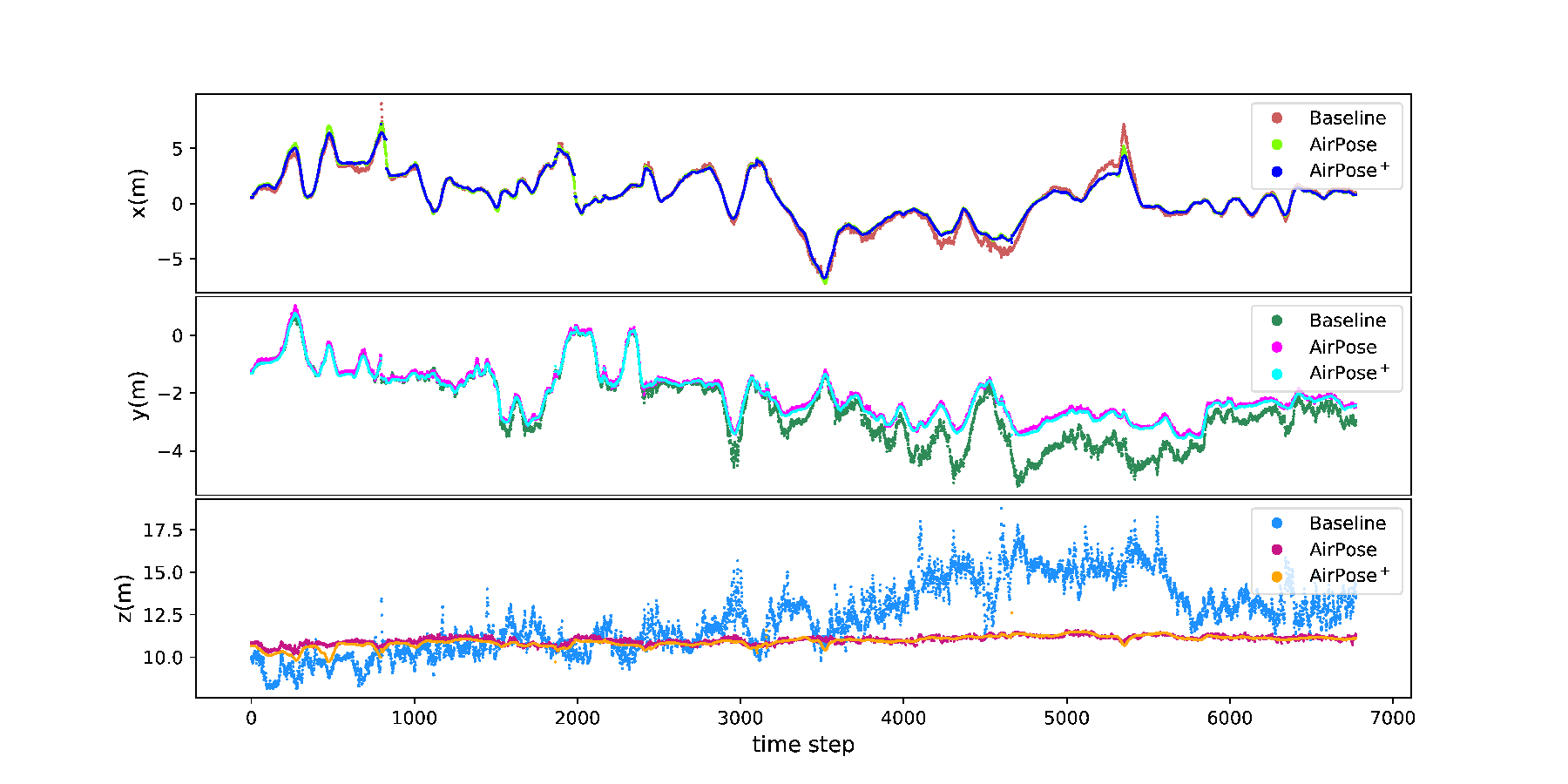}   
  \end{tabular}
  \caption{Estimated position of person 1 by the baseline method, AirPose and AirPose+ relative to UAV1 (left) and UAV2 (right).}
\label{fig:person1_traj}  
\end{figure*}

\noindent \textbf{AirPose:} After training on the synthetic data, we fine-tune the `Baseline' and the proposed method, AirPose, on the real image data sequence of person 1 (and test on person 2) using a hardware-in-the-loop setup as described in Sec.~\ref{sec:training}. Since we do not have the camera extrinsic parameters (the global position of the DJI UAV is not accessible), we cannot transform the estimated SMPL-X parameters into a global reference frame to calculate any quantitative metric for comparison, as we did in the synthetic case. 
However, for both baseline and AirPose, we show the plots of the X, Y and Z coordinates of the estimated position w.r.t.\ the UAV cameras in Fig.~\ref{fig:cam1_trajectory} of person 2 and Fig.~\ref{fig:person1_traj} of person 1. 
The results in these plots demonstrate that the position estimate of the `Baseline' method is significantly noisier than that of AirPose. 
In particular, the estimate in the Z direction (i.e.~depth estimate from the camera) for UAV 2 has unrealistic variations when using the `Baseline' method.
In contrast, AirPose estimates a much more realistic depth and a smoother, more realistic trajectory of the person. In Fig.~\ref{fig:spinRes}, we compare the depth estimate of our method with vanilla SPIN. The vanilla SPIN depth estimate is significantly noisier, similar to the baseline method, because in both cases the estimation is done on the cropped and scaled image.



\begin{figure*}[!htbp]
\setlength{\abovecaptionskip}{-5pt} 
      \centering
      \begin{tabular}{cc}
      \hspace*{-20pt} \includegraphics[width=0.53\linewidth]{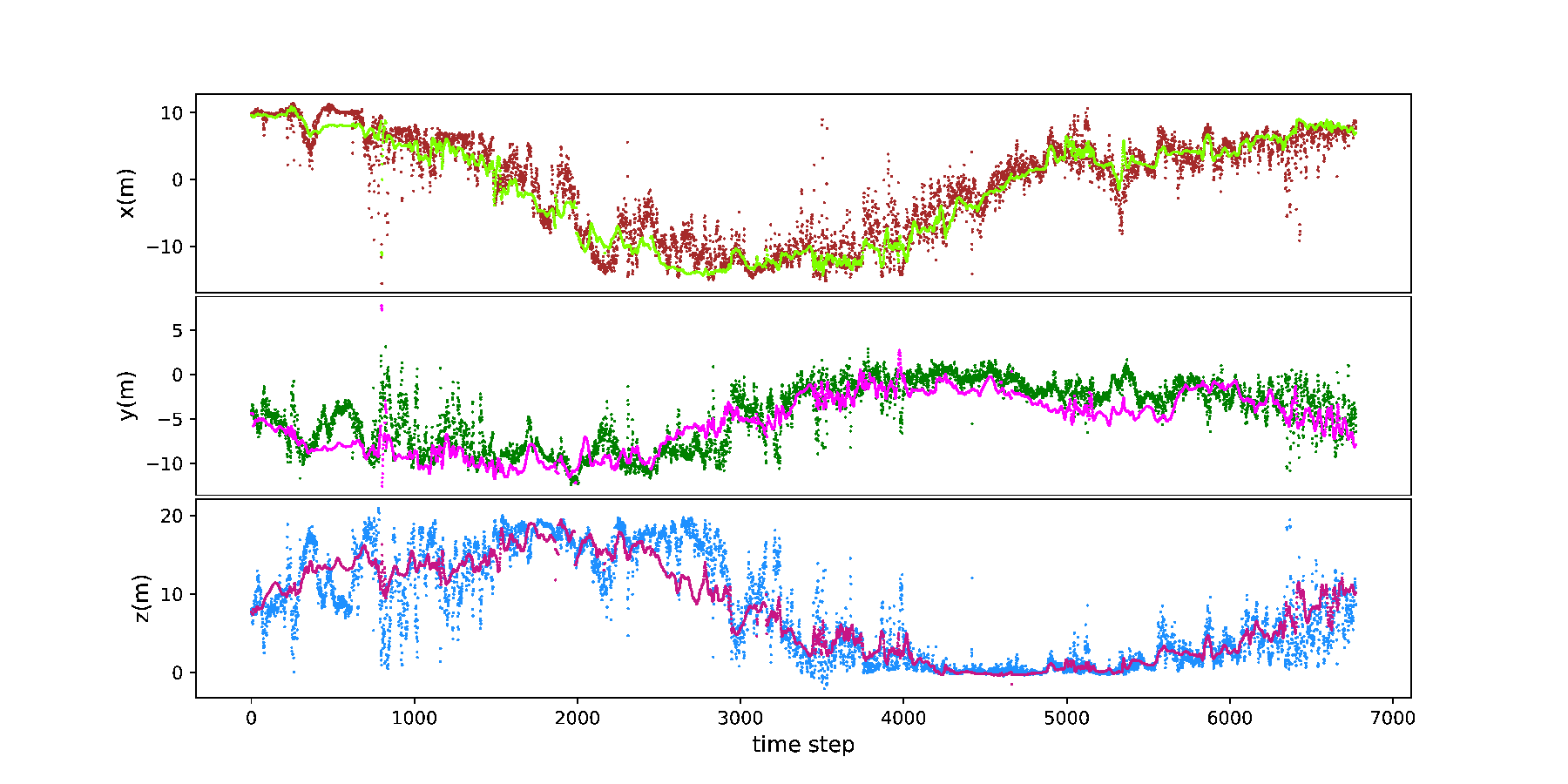} & \hspace*{-40pt} \includegraphics[width=0.53\linewidth]{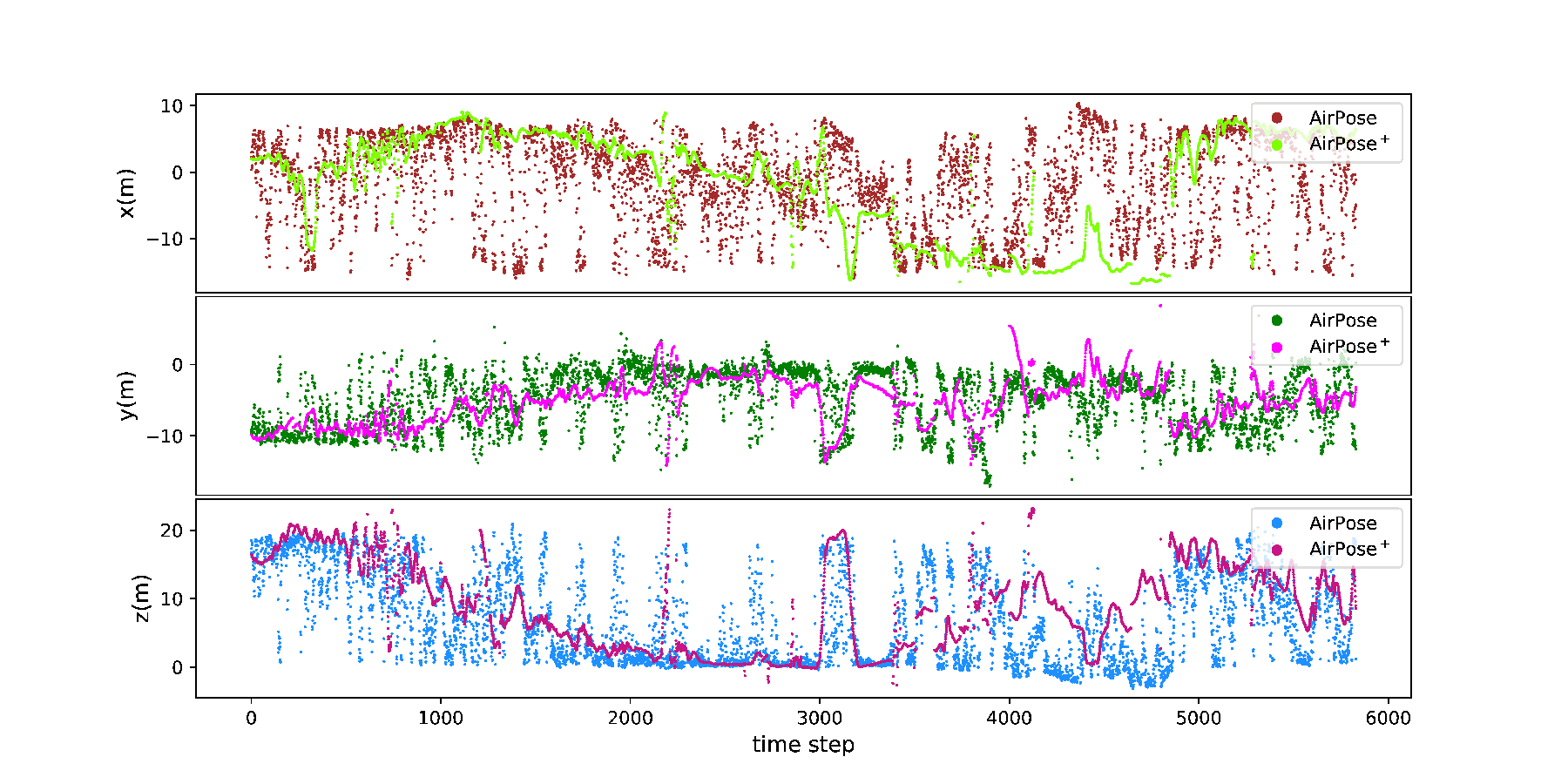}
      \end{tabular}
    \caption{Estimated trajectory of UAV 2 w.r.t.~UAV1 for the capture sequence with person 1 (left) and 2 (right). AirPose: brown (x), green (y) and blue (z). AirPose$^+$: light green (x), pink (y) and magenta (z).}
    \label{fig:UAV2_trajectoryS2}
\end{figure*}

\noindent \textbf{AirPose$^+$:} We further show the results of AirPose and AirPose$^+$ on the real data (see also the attached video).
Figure~\ref{fig:real_data_results} shows the estimated mesh overlaid on the images. We can see AirPose$^+$ improves the AirPose estimation of the articulated pose.   
Figure~\ref{fig:UAV2_trajectoryS2} shows the estimated position of UAV 2 w.r.t.~UAV 1, which was kept hovering in place. We use the estimated SMPL-X pose in each UAV frame to calculate the position of UAV 2 w.r.t.~UAV 1. This position estimate is extremely sensitive to even minor estimation errors in any of the two views. A small error in the SMPL-X rotation estimate leads to a significant error in UAV 2's position. For person 1, we can see that the UAV 2 trajectory  estimated by AirPose is a bit noisy. AirPose$^+$, however, improves this estimate further, as seen in Fig.~\ref{fig:UAV2_trajectoryS2} (left). 

The sequence with person 2 contains several complex twisting poses in which the upper body rotates w.r.t.~the lower body. For such poses, AirPose fails to estimate the correct the root rotation of the person. This results in the estimated UAV 2 position by AirPose being very noisy, as seen in Fig.~\ref{fig:UAV2_trajectoryS2} (right). However, the position estimate of UAV 2 is significantly improved in this case by AirPose$^+$. Nevertheless, it still has a few sudden jumps. In addition to the twisting poses, errors in keypoint detection are also responsible for errors in the person 2's sequence. Such errors are common with 2D detectors, e.g., left/right swap of 2D keypoints.


\section{Conclusions}
\label{sec:conclusion}
We have presented a new approach for human 3D pose and shape estimation using multiple UAVs. Our novel network architecture is decentralized, distributed, light-weight and requires little inter-UAV communication, making it suitable for on-board deployment on UAVs. We demonstrated that our approach successfully fuses information from multiple viewpoints, significantly improving pose estimates of both the person and the UAVs relative to the person, when compared to baseline methods.
\
We introduced a powerful procedure to train such a network using large computer-generated datasets of synthetic images in virtual environments, and to fine-tune on a small set of real images. 
\
Through a systematic evaluation on synthetic data, we show that AirPose is significantly more accurate than the state-of-the-art method adapted for this problem. 
\
On real image data, captured by two UAVs, we show substantial qualitative improvement over the state-of-the-art method. 
\
Thus, AirPose overcomes significant problems currently limiting the deployment of aerial MoCap systems in areas such as search and rescue and aerial cinematography.


%

{
	\bibliographystyle{IEEEtran}
	\bibliography{AirPose}
}

\end{document}